\documentclass{article} % For LaTeX2e
\usepackage{iclr2016_conference,times}
\usepackage{hyperref}
\usepackage{latexsym}
\usepackage{amsmath}
\usepackage{multirow}
\usepackage{url}
\usepackage[utf8]{inputenc}
\usepackage{graphicx}
\usepackage[encapsulated]{CJK}
\usepackage{ucs}
\usepackage{rotating}
\usepackage{algorithmic}
\usepackage{algorithm}
\usepackage{color}
\usepackage{amssymb}

\newcommand{\ignore}[1]{}

\newcommand{\examp}[1]{\emph{#1}} 

\title{Character-based\\
Neural Machine Translation}

\author{Wang Ling \& Isabel Trancoso\\
L$^2$F Spoken Systems Lab\\
Instituto Superior T\'{e}cnico\\
Lisbon, Portugal \\
\texttt{\{wlin,isabel.trancoso\}@inesc-id.pt} \\
\AND
Chris Dyer \& Alan Black \\
Language Technologies Institute \\
Carnegie Mellon University \\
Pittsburga, PA 15213, USA \\
\texttt{\{cdyer,awb\}@cs.cmu.edu} \\
}

\begin{document}

\maketitle

\begin{abstract}
We introduce a neural machine translation model that views the input and output sentences as sequences of characters rather than words. Since word-level information provides a crucial source of bias, our input model composes representations of character sequences into representations of words (as determined by whitespace boundaries), and then these are translated using a joint attention/translation model. In the target language, the translation is modeled as a sequence of word vectors, but each word is generated one character at a time, conditional on the previous character generations in each word. As the representation and generation of words is performed at the character level, our model is capable of interpreting and generating unseen word forms. A secondary benefit of this approach is that it alleviates much of the challenges associated with preprocessing/tokenization of the source and target languages. We show that our model can achieve translation results that are on par with conventional word-based models.
\end{abstract}

\section{Introduction}

In the past, efforts at performing translation at the character-level~\citep{Vilar:2007:WTL:1626355.1626360} or subword-level~\citep{neubig13mtj} have failed to produce competitive results compared to word-based counterparts~\citep{Brown:1993:MSM:972470.972474,moses,Chiang:2005:HPM:1219840.1219873}, with the exception of closely related languages~\citep{DBLP:conf/acl/NakovT12}. However, developing sequence to sequence models that are capable at reading and generating and generating words at the character basis is attractive for multiple reasons. Firstly, it opens the possibility for models reason about unseen source words, such as morphological variants of observed words. Secondly, it allows the production of unseen target words effectively recasting translation as an open vocabulary task. Finally, we benefit from a significant reduction of the source and target vocabulary size as only characters need to be modelled explicitly. As the number of word types increases rapidly with the size of the dataset~\cite{heaps:1978}, while the number of letter types in the majority languages is fixed, character MT models can potentially solve many scalability issues in MT, both in terms of computational speed and memory requirements. Namely, the computational cost of performing a softmax over the whole vocabulary, and the memory needed to represent each existing word type explicitly.

In this work, we present a neural translation model that learns to encode and decode using at the character level. We show that contrarily to previous belief~\citep{Vilar:2007:WTL:1626355.1626360,neubig13mtj}, models that work at the character level can generate results competitive with word-based models. This is accomplished by indirectly incorporating the knowledge of words into the model using a hierarchical architecture that generates the word representations from characters, maps the word representation into the target language and the continuous space and then proceeds to generate the target word character by character. As the composition of the words is based on characters, the model can learn, for instance, morphological aspects in the source language, allowing it to build effective representations for unseen words. On the other hand, the character-based generation model allows the model to perform translation into word forms that are not present in the training corpus. 

\section{Character-based Machine Translation Model}
\label{sec:c2v}

This section describes our character-based machine translation model. As an automatic translation task, it perform the translation of a source sentence $\boldsymbol{s}=s_0,\ldots,s_n$, where $s_i$ is the source word at index $i$ into the target sentence $\boldsymbol{t}=t_0,\ldots,t_m$, where $t_j$ is the target word at index $j$. This can be decomposed into a serie of predictions, where the model predicts the next target word $t_p$, given the source sentence $\boldsymbol{s}$ and the previously generated target words $t_0,\ldots,t_{p-1}$. 

\paragraph{Notation} We shall represent vectors with lowercase bold letters (e.g. $\mathbf{a}$), matrixes with uppercase bold letters (e.g. $\mathbf{A}$), scalar values as regular lowercase letters (e.g. $a$) and sets as regular uppercased letters (e.g. $A$). When referring to whole source and target sentences, we shall use the variables $s$ and $t$, respectively. Individual source and target words, shall be referred as $s_i$ and $t_j$, where $i$ and $j$ are the indexes of the words within the sentence. Furthermore, we use variables $n$ and $m$ to refer to the lengths of the source and target sentences. Finally, we shall define that $s_0$ and $t_0$ represent a special start of sentence token denoted as \examp{SOS} and that $s_n$ and $t_m$ represent a special end of sentence token denoted as \examp{EOS}. To refer to individual characters, we shall use the notation $s_{i,u}$ and $t_{j,v}$, which denotes the $u$-th character in the $i$-th word in the source sentence and to the $v$-th character in the $j$-th word in the target sentence, respectively. We use the variables $x$ and $y$ as the lengths of the source and target words. We also define that the first character within a word is always a start of word token denoted as \examp{SOW} and the last characters $s_{i,x}$ and $s_{j,y}$ are always the end of word character \examp{EOW}. 

\subsection{Joint Alignment and Translation Model}

\begin{figure*}[ht]
\begin{center}
\centerline{\includegraphics[width=0.9\columnwidth]{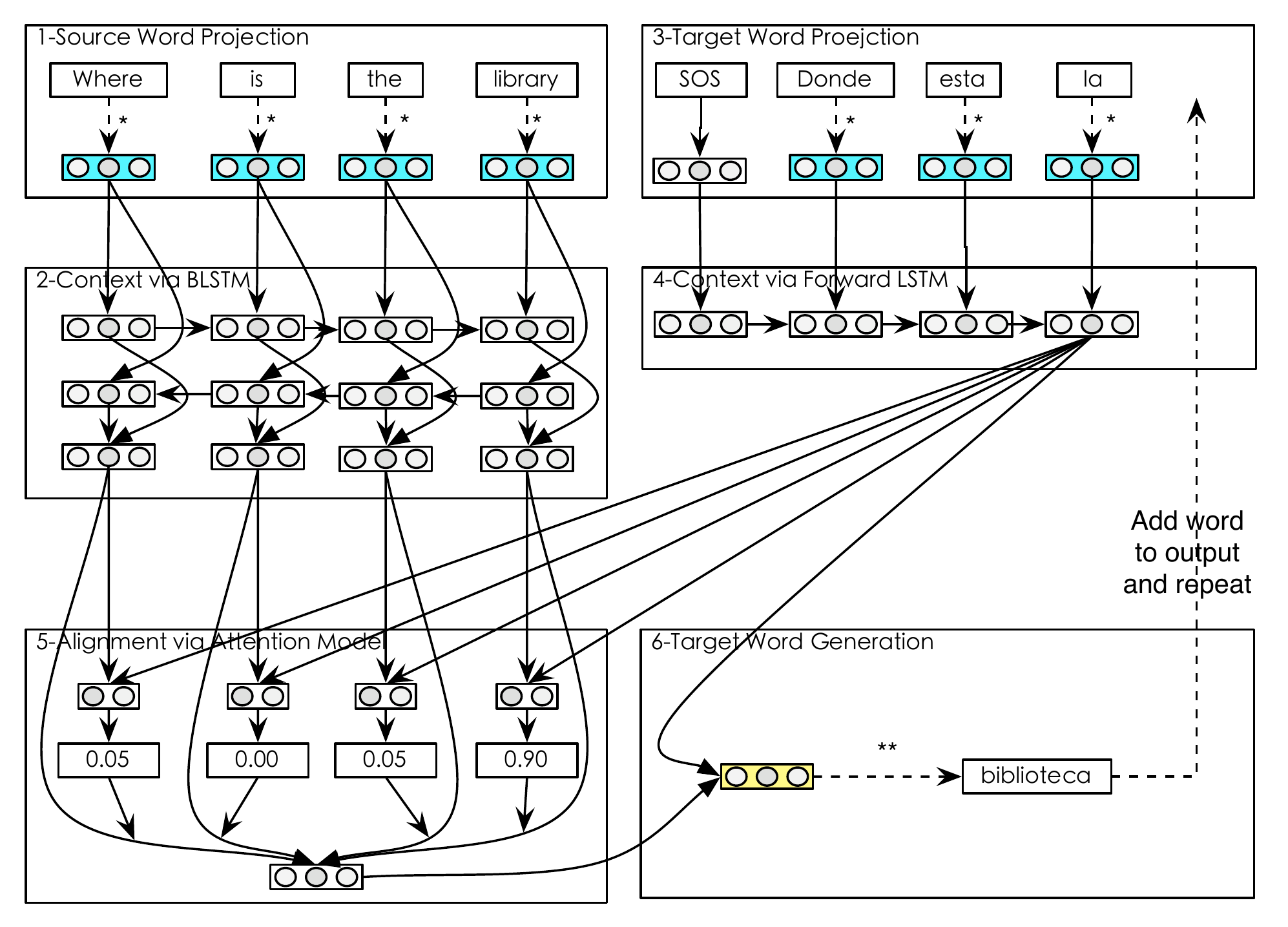}}
\caption{Illustration of the joint alignment and translation model. Square boxes represent vectors of neuron activations.}
\label{mtmodel}
\end{center}
\end{figure*}

While our model can be applied to any neural translation model , we shall adapt the attention-based translation model presented in~\citep{DBLP:journals/corr/BahdanauCB14}, which is described as follows. An illustration of the model is shown in Figure~\ref{mtmodel}. In this model, the translation of a new target word $t_p$, given the source sentence $s_0,\ldots,s_n$ and the current target context $t_0,\ldots,t_{p-1}$ is performed in the following steps. 

\paragraph{1-Source Word Projection} Source words are mapped into $d_{s,w}$-dimentional vectors. The most common approach to perform this projection is through a word lookup table, where each word type is attributed a independent set of parameters. Thus, a word lookup table with $S$ words will require $d_{s,w}\times S$ parameters, where $d_{s,w}$ is the size of the word embeddings.

\paragraph{2-Context via BLSTMs}
%The information contained within each projected source word $\mathbf{s}_0,...,\mathbf{s}_n$, does not hold any information regarding their position in the sentence, which is limitative for many reasons (e.g. determining the sense of ambiguous words). While there are many ways for encoding this information, such as convolutional and window-based approaches~\citep{collobert2011natural}, a popular approach is to use recurrent neural networks to encode global context. We use Long-Short Memory RNNs (LSTMs), which are similar to recurrent neural networks in that given a set of input vectors $\mathbf{i}_0,...,\mathbf{i}_k$, it produces a set of states $\mathbf{h}_0,...,\mathbf{h}_k$, where each state $\mathbf{h}_j$ is the result of the composition between the current input $\mathbf{i}$ and the previous state $\mathbf{h}_{j-1}$. The main difference between RNNs and LSTMs is that a cell is added at each timestamp $j$, which is updated linearly, so that it is not affected by the vanishing gradient effect in non-linear updates. This allows LSTMs to learn long range dependencies more effectively than RNNs. We shall not detail the inner workings of LSTMs (These can be found in~\citep{Hochreiter:1997:LSM:1246443.1246450}), as from the input to output perspective both RNNs and LSTMs produce a state sequence from an input sequence. 

A context-aware representation of each source word vector $\mathbf{s}_0,\ldots,\mathbf{s}_n$ is obtained by using two LSTMs~\citep{Hochreiter:1997:LSM:1246443.1246450}. The forward LSTM generates the state sequence $\mathbf{g}^f_0,\ldots,\mathbf{g}^f_{n}$ from the input sequence $\mathbf{s}_0,\ldots,\mathbf{s}_n$, and encode the left context. Then, the backward LSTM, reads the input sequence in the reverse order $\mathbf{s}_n,\ldots,\mathbf{s}_0$ generating the backward state sequence $\mathbf{g}^b_n,\ldots,\mathbf{g}^b_{0}$, which encodes the right context. Consequently, for each word $s_i$, the global context is obtained as a linear combination of the respective forward state $\mathbf{g}^f_i$ and backward state $\mathbf{g}^b_i$.

\paragraph{3-Target Word Projection}
The projection of target words essentially follows the same approach as the projection of the source words. A distinct word lookup table for the target language is used with $d_{t,w}\times T$ parameters, where $d_{t,w}$ is the dimensionality of the target word vectors and $T$ is the size of the target vocabulary.

\paragraph{4-Context via Forward LSTM}
Unlike the source sentence $s$, the target sentence is not known a priori. Thus, a forward LSTM is built over the translated sentence $t_0,\ldots,t_{p-1}$, generating states $\mathbf{l}^f_0,\ldots,\mathbf{l}^f_{p-1}$. As result, state $\mathbf{l}^f_{p-1}$ encodes the currently translated context.  %Similarly to a language model, for the prediction of word $t_p$, we use state $\mathbf{l}^f_{p-1}$, which encodes the left context prior to the prediction of $t_p$. %This model is simply a LSTM, which is hyper-parametrized by the state and cell sizes, $l_{state}$ and $l_{size}$, and defines the update parameters $L_{lstm}$. As input, the model receives the currently translated target words $t_0,...,t_{p-1}$ and returns the last state of the model encoding the current target context $\mathbf{l}^f_{p-1}$.

\paragraph{5-Alignment via Attention}

For each target context $\mathbf{l}^f_{p-1}$, the attention model learns the attention that is attributed to each of the source vectors $\mathbf{b}_0,\ldots,\mathbf{b}_n$. Essentially, the model computes a score $z_i$ for each source word by applying the following function:

\begin{align*}
z_i = \mathbf{s} \tanh(\mathbf{W}_t \mathbf{l}^f_{p-1} + \mathbf{W}_s \mathbf{b}_i),
\end{align*}

where the source vector $\mathbf{b}_i$ and target vector $\mathbf{l}^f_{p-1}$ are combined into a $d_s$-dimensional vector using the parameters $\mathbf{W}_s$ and $\mathbf{W}_t$, respectively, and a non-linearity is applied (In our case, we apply a hiperbolic tangent function $\tanh$). Then, the the score is obtained using the vertical vector $\mathbf{s}$. This operation is performed for each source index obtaining the scores $z_0,\ldots,z_n$. Then, a softmax over all scores as is performed as follows:

\begin{align*}
a_i = \frac{\exp(z_i)}{\sum_{j\in[0,n]} \exp(z_j)}
\end{align*}

This function yields a set of attention coefficients $a_0,\ldots,a_n$ with the property that each coefficient is a value between 0 and 1, and the sum of all coefficients is 1. In MT, this can be interpreted as a soft alignment for the target word $t_p$ over each of the source words. These attentions are used to obtain representation of the source sentence for predicting word $w_p$, which is simply the weighted average $\mathbf{a} = \sum_{i\in[0,n]} a_i \mathbf{b}_i$.

%Thus, the attention model defines the hyper-parameter $d_{z}$, which is the size of the hidden layer that combines $\mathbf{l}^f_{p-1}$ and each $\mathbf{b}_i$. Then, the parameters $\mathbf{W}_s$ and $\mathbf{W}_t$ project the input vectors into this hidden layer and the vertical vector $s$ that scores this hidden layer. The attention model, takes as input the target context vector $\mathbf{l}^f_{p-1}$ and a set of source vectors $\mathbf{b}_0,\ldots,\mathbf{b}_n$, and returns a vector $\mathbf{a}$, which is the average of all source vectors $\mathbf{b}_0,\ldots,\mathbf{b}_n$, weighted by their estimated attentions $a_0,\ldots,a_n$.

\paragraph{6-Target Word Generation}
In the previous steps, the model builds two vectors that are used for the prediction of the next word in the translation $t_p$. The first is the target context vector $\mathbf{l}^f_{p-1}$, which encodes the information of all words preceding $t_p$, and $\mathbf{a}$, which contains the information of the most likely source word/words to generate $w_p$. A common approach to word prediction is to apply a softmax function over the target language vocabulary. More formally, given the conditioned variables, the source attention $\mathbf{a}$ and the target context $\mathbf{l}^f_{p-1}$, the probability of a given word type $t^p$ being the next translated word $t_p$ is given by:

\begin{align*}
P(t^p|\mathbf{a},\mathbf{l}^f_{p-1}) = \frac{\exp(e^{\mathbf{S}^{t^p}_a\mathbf{a} + \mathbf{S}^{t^p}_l \mathbf{l}^f_{p-1}})}{\sum_{j\in[0,T]} \exp(e^{\mathbf{S}^j_a\mathbf{a} + \mathbf{S}^j_l \mathbf{l}^f_{p-1}})},
\end{align*}

where $\mathbf{S}_a$ and $\mathbf{S}_l$ are the parameters that map the conditioned vectors into a score for each word type in the target language vocabulary $T$. The parameters for a specific word type $j$ are obtained as $\mathbf{S}^j_a$ and $\mathbf{S}^j_l$, respectively. Then, scores are normalized into a probability.

\subsection{Character-based Machine Translation}

We now present our adaptation of the word-based neural network model to operate over character sequences rather than word sequences. However, unlike previous approaches that attempt to discard the notion of words completely~\citep{Vilar:2007:WTL:1626355.1626360,neubig13mtj}, we propose an hierarhical architecture, which replaces the word lookup tables (steps 1 and 3) and the word softmax (step 6) with character-based alternatives, which compose the notion of words from individual characters. The advantage of this approach is that we benefit from properties of character-based approaches (e.g. compactness and orthographic sensitivity), but can also easily be incorporated into any word-based neural approaches.

\paragraph{Character-based Word Representation}

The work in~\citep{ling:2015,miguel:2015} proposes a compositional model for learning word vectors from characters. Similar to word lookup tables, a word string $s_j$ is mapped into a $d_{s,w}$-dimensional vector, but rather than allocating parameters for each individual word type, the word vector $\mathbf{s}_j$ is composed by a series of transformation using its character sequence $s_{j,0},\ldots,s_{j,x}$. 

\begin{figure}[ht]
%\vskip -0.2in
\begin{center}
\centerline{\includegraphics[width=1.0\columnwidth]{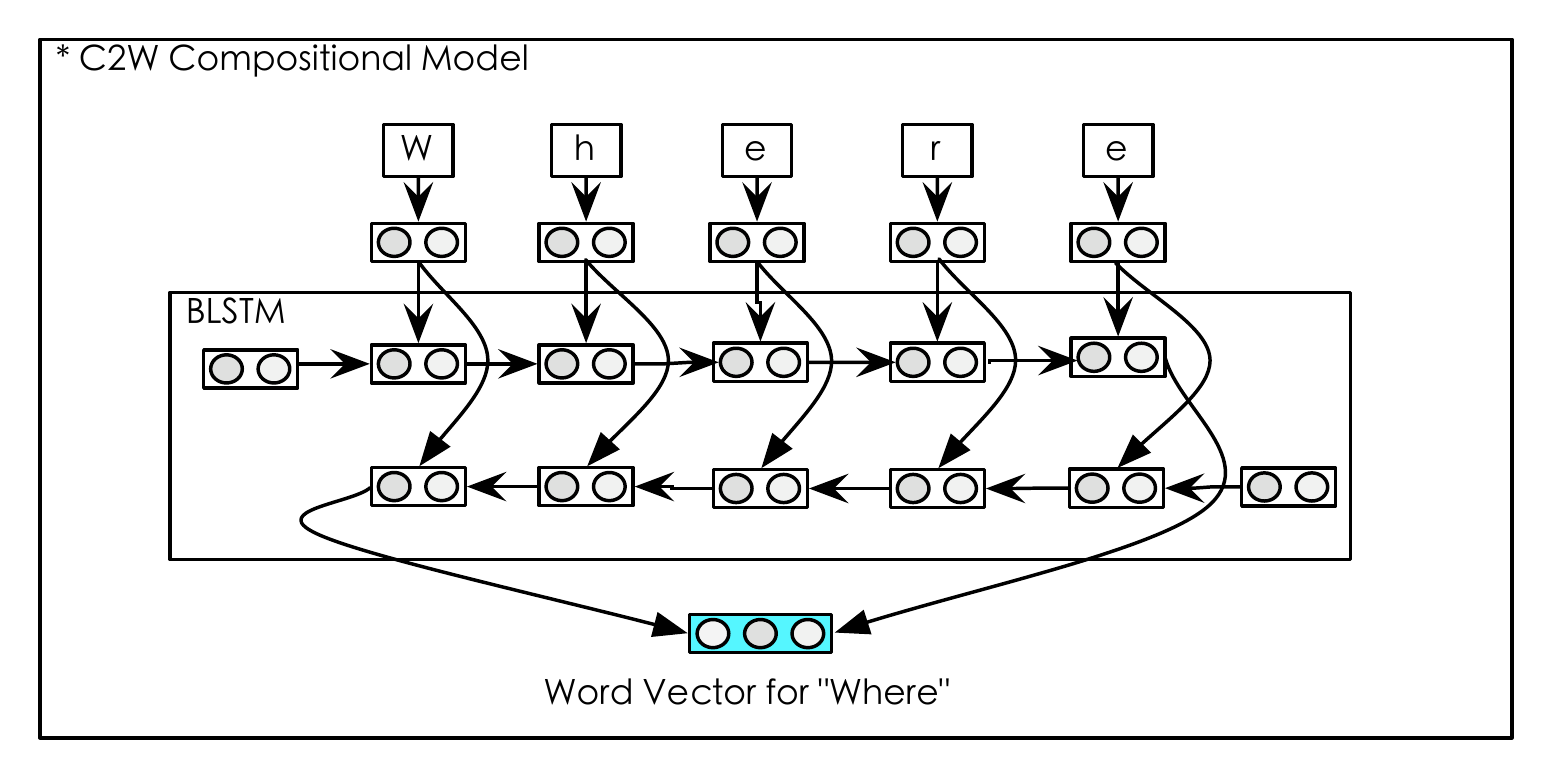}}
\caption{Illustration of the C2W model. Square boxes represent vectors of neuron activations.}
\label{c2w}
\end{center}
%\vskip -0.4in
\end{figure} 

The illustration of the model is shown in~\ref{c2w}. Essentially, the model builds a representation of the word using characters, by reading characters from left to right and vice-versa. More formally, given an input word $s_{j}=s_{j,0},\ldots,s_{j,x}$, the model projects each character into a continuous $d_{s,c}$-dimensional vectors $\mathbf{s}_{j,0},\ldots,\mathbf{s}_{j,x}$ using a character lookup table. Then, it builds a forward LSTM state sequence $\mathbf{h}^f_0,\ldots,\mathbf{h}^f_{k}$ by reading the character vectors $\mathbf{s}_{j,0},\ldots,\mathbf{s}_{j,x}$. Another, backward LSTM reads the character vectors in the reverse order generating the backward states $\mathbf{h}^b_{k},\ldots,\mathbf{h}^b_0$. Finally, the representation of the word $\mathbf{s}_j$ is obtained by combining the final states as follows:
\begin{align*}
\mathbf{s}_j = \mathbf{D}_{s,f} \mathbf{h}^f_k + \mathbf{D}_{s,b} \mathbf{h}^b_0 + \mathbf{b}_{s,d},
\end{align*}
where $\mathbf{D}_{s,f}$, $\mathbf{D}_{s,b}$ and $\mathbf{b}_{s,d}$ are parameters that determine how the states are combined. 

As the C2W model maps a word string into a vector, we can simply replace the word lookup tables (steps 1 and 3) with two C2W models in order to obtain a character-based translation model at the input. Next, we shall describe a method to output words at the character level.

%As hyperparameters, the C2W model also defines $d_{s,w}$ and $d_{t,w}$ as the size of the generated word vectors for the source and target language. Then, it defines $d_{s,c}$ and $d_{t,c}$ as the size of the character vectors. Finally, it defines the size of the LSTM states $d_{s,state}$, $d_{t,state},d_{s,cell}$, $d_{t,cell}$. As parameters, we define the BLSTM paramaters as $D_{s,blstm}$ and $D_{t,blstm}$ and the parameters governing the combination of the BLSTM final states $\mathbf{D}_{s,f}$, $\mathbf{D}_{s,b}$, $\mathbf{b}_{s,d}$, $\mathbf{D}_{t,f}$, $\mathbf{D}_{t,b}$ and $\mathbf{b}_{t,d}$.

\paragraph{Character-based Word Generation}
A word softmax requires a separate set of parameters for each word type. Thus, a word softmax cannot generate unseen words in the training set, and requires a large amount of parameters due to the fact that each word type must be modelled independently. Furthermore, another well know problem is that the whole target vocabulary $T$ must be traversed for each prediction during both training and testing phases. While at training time approximations such as noise contrastive estimation~\citep{GutmannHyvarinen:NoiseContrastiveEstimation} can be applied, traversing $T$ is still required at test time. 

\begin{figure}[ht]
%\vskip -0.2in
\begin{center}
\centerline{\includegraphics[width=1.0\columnwidth]{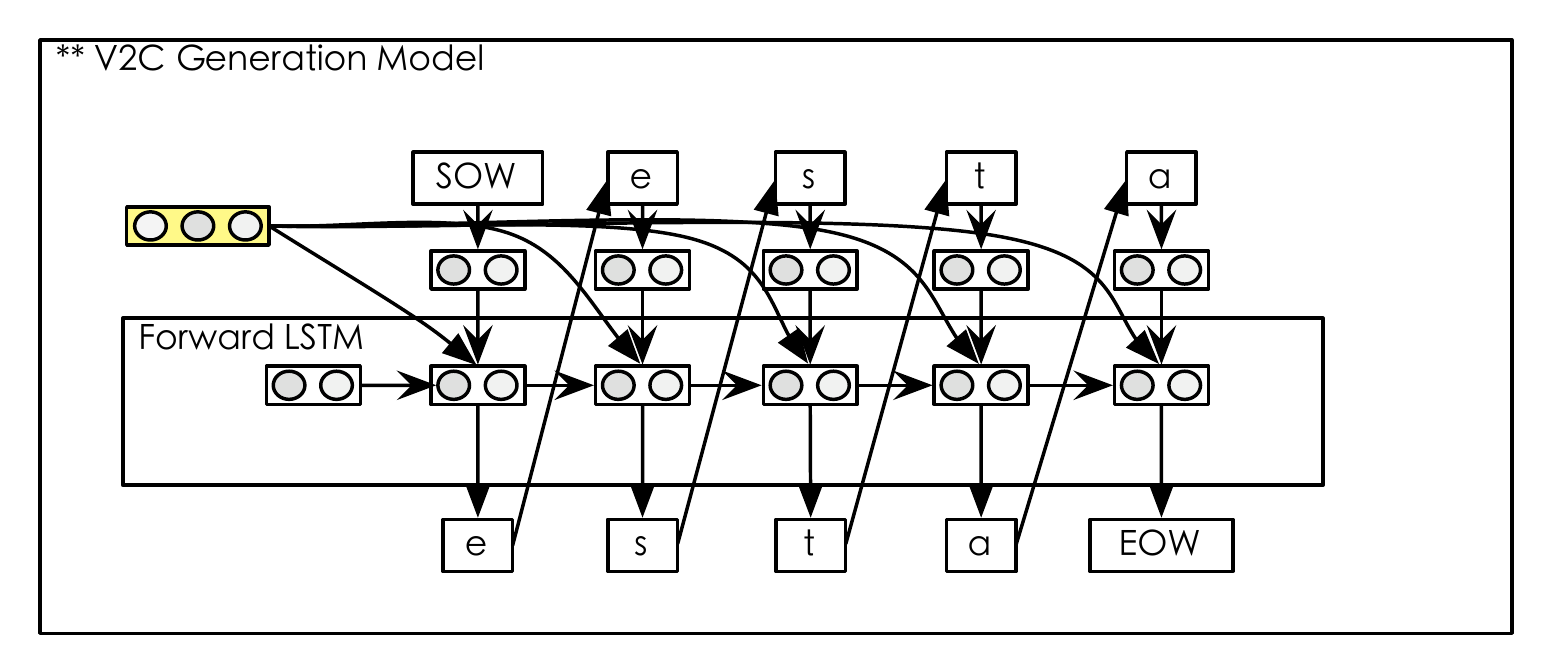}}
\caption{Illustration of the V2C model. Square boxes represent vectors of neuron activations.}
\label{v2c}
\end{center}
%\vskip -0.4in
\end{figure} 

We address these problems by defining a character-based word generation model. An illustration of the V2C (vector to characters) is shown in Figure~\ref{v2c}. We define a character vocabulary for the target language $T_c$ and a given word $t^{j}$ as a sequence of characters $t^{j,0},\ldots,t^{j,y}$, the probability of a given word is redefined as:

\begin{align*}
P(w_p|\mathbf{a},\mathbf{l}^f_{p-1}) = \prod_{i\in [0,y]} P(w_{j,i}|w_{k,0},\ldots,w_{j,j-1},\mathbf{a},\mathbf{l}^f_{p-1})
\end{align*}

The intuition is that rather than learning to predict single words, our model predicts the character sequence of the output word. Each prediction is dependent on the input of the model (aligned source words $\mathbf{a}$ and and target word context $\mathbf{l}^f_{p-1}$) and also on the previously generated character context $t_{j,0},\ldots,t_{j,q-1}$, where $q$ is the index of the last predicted character.

The character context is generated by a LSTM. First, we project each target character $t_{j,0},\ldots,t^{j,q-1}$ with a character lookup table in the target language (we use the same lookup table used in the C2W model for the target language) into a $d_{t,c}$-dimensional vector $\mathbf{t}_{j,0},\ldots,\mathbf{t}_{j,q-1}$. Then, each vector is concatenated to the vectors $\mathbf{a}$ and $\mathbf{l}^f_{p-1}$, and passed as the input to an LSTM, generating the sequence of states $\mathbf{y}^f_0,\ldots,\mathbf{y}^f_{q-1}$. Then, the prediction of the character $t_{k,q}$ obtained as the softmax function:

\begin{align*}
P(t_{j,q}|t_{j,0},\ldots,t_{j,q-1},\mathbf{a},\mathbf{l}^f_{p-1}) = \frac{\exp(S^{w_{k,q}}_{y} \mathbf{y}^f_{q-1})}{\sum_{i\in[0,T_c]} \exp(S^i_y\mathbf{y}^f_{q-1})},
\end{align*}

where $S_y$ are the parameters that convert the state $y$ into a score for each output character, and $S^i_y$ denotes the parameters respective to the character type $i$. The word terminates once the end of word token \examp{EOW} is generated.

Finally, the model is also required to produce the end of sentence token \examp{EOS}, similarly to a word softmax. In our model, we simply consider the \examp{EOS} token as a word whose only character is \examp{EOS}. In another words, it must generate the sequence \examp{EOS},\examp{EOS}. 

%The hyper parameters of our model, are essentially the LSTM parameters (cell and state size) $y_{cell}$ and $y_{state}$. As for the paramemters, the V2W model requires an LSTM $\mathbf{Y}_{LSTM}$ and the parameters for the character softmax $S_{y}$. 

\subsection{Training}
During training the whole set of target words $t_0,\ldots,t_m$ are known, and we simply maximize the log likelihood that the sequence of words $\log p(t_{0} \mid \mathbf{a},\textit{SOS}) + \ldots + \log p(t_{m} \mid\mathbf{a},\mathbf{l}^f_{m-1})$. More formally, we wish to maximize the log likelihood of the training data defined as:
\begin{align*}
\sum_{(\boldsymbol{s},\boldsymbol{t})\in D} \sum_{p\in [0,m]} \log p(t_{q} \mid \mathbf{a},\mathbf{l}^f_{m-1})
\end{align*}
where $D$ is the set of parallel sentences used as training data. 

In the word-based model, optimizing this function can be performed by maximizing the word softmax objective. In the character-based model, this each word prediction is factored into a set of character predictions. More concretely, we we maximize the following log-likelihood:

\begin{align*}
\sum_{(\boldsymbol{s},\boldsymbol{t})\in D} \sum_{p\in [0,m]} \sum_{q\in [0,length(t_p)]} \log p(t_{p,q} \mid	 t_{k,0},\ldots,t_{k,q-1},\mathbf{a},\mathbf{l}^f_{m-1})
\end{align*}

\subsection{Decoding}
In previous work~\citep{DBLP:journals/corr/BahdanauCB14}, decoding is performed using beam search. In the word-based approach, we define a stack of translation hypothesis per timestamp $\mathbf{A}$, where each position is a set of translation hypothesis. With $\mathbf{A}_0 = SOS$, at each timestamp $j$, we condition on each of the previous contexts $t=\mathbf{A}_{j-1}$ and add new hypothesis for all words obtained in the softmax function. For instance, given the partial translations $\mathbf{A}_1 = {A,B}$, and the vocabulary $T={A,B,C}$, then $\mathbf{A}_2$ would be composed by ${AA, AB, AC, BA, BB, BC}$. We set a beam $k_w$, which defines the number of hypothesis to be expanded prioritizing hypothesis with the highest sentence probability. An hypothesis is final once it generates the end of sentence token \examp{EOS}.

Whereas, the word softmax simply returns a list of candidates for the next word by iterating through each of the target word types, the V2C model can generate an infinite number of target words. Thus, in the character-based MT model, a second decoding process is needed to return the list of top scoring words at each timestamp. That is, we define a second beam search decoder with beam $k_c$, and perform a stack-based coding on the character-level for each word prediction. The decoder defines a stack $\mathbf{B}$, where at each timestamp, a new character is generated for each hypothesis. In this case, the beam search is run until $k_w$ final hypothesis are found (generation of EOW), as it must return at least $k_w$ new hypothesis to ensure that the word level search is complete.

\subsection{Layer-wise Training}
Our character-based model defines a three layer hierarchy. Firstly, characters are composed into word vectors using the C2W model. Then, the attention model searches for the next source word to translate. Finally, the generation of the target word is obtained using the V2C model. Each of these layers contain in many cases multiple non-linear projections, training is bound to be significantly more complex than word-based models. In practice, this causes more epochs to be necessary for convergence, and the model can converge to a suboptimal local optimum. Furthermore, while the V2C model is generally more efficient than a word softmax, the introduction of the C2W model significantly slows down training, as simple word table lookup is replaced by a compositional model. 

Inspired by previous work on training deep multi-layered perceptrons~\citep{NIPS2006_3048}, we start by training the attention and V2C models, which are directly linked to the output of the model. The C2W model, which is the bottleneck of the model, is temporarily replaced by word lookup tables, and the model is trained to maximize the translation score in the development set. 

The C2W model is introduced afterwards by first training the C2W model to produce the same word vectors as the word lookup tables for all training word types. More formally, given a word $w$, and the embeddings from the word lookup table $\hat{\mathbf{w_1}},\ldots,\hat{\mathbf{w_{d_w}}}$, and the embeddings produced by the C2W model $\mathbf{w_1},\ldots,\mathbf{w_{d_w}}$, we wish to optimize the parameters to minimize the square distance $(\hat{\mathbf{w_1}}-\mathbf{w_1})^2,\ldots,(\mathbf{w_{d_w}})^2$. As result, C2W model will generate similar embeddings as the word lookup table, and replacing them will not degenerate the results significantly. Finally, the full set of parameters (C2W, attention model and V2C) are fine-tuned to maximize the translation quality on the development set.

\subsection{Weak Supervision for Attention Model}
A problem with attention models is the fact that finding the latent attention coefficients $a_i$, requires a large number of epochs~\citep{nips15_hermann}. Furthermore, as the attention model defined in~\citep{DBLP:journals/corr/BahdanauCB14} does not define any domain knowledge regarding word alignments, such as distortion, symmetry and fertility~\citep{Brown:1993:MSM:972470.972474} this model is likely to overfit for small amounts of data. To address this problem, we use IBM model 4 to produce the word alignments for the training parallel sentences, impose a soft restriction to induce the attention model to produce alignments similar to word alignments produced by the IBM model. More formally, given that the target word is aligned to the source word at index $k$ in IBM model 4, we wish the model to maximize the coefficient $a_k$. As $a_k$ is obtained from a softmax, this is essentially means that we wish to maximize the probability that the target word $w_k$ is selected. As word alignments tend to be one-to-one, for target words with multiple or no alignments, we simply set no soft restriction. 

\section{Experiments}
In this section, we present and analyse the results using our proposed character-based model.

\subsection{Setup}
We test our model in two datasets. First, we 600k sentence pairs for training from Europarl~\citep{koehn2005epc}, in the English-Portuguese language pair. Then, we define another 500 sentence pairs for development and 500 sentence pairs for testing. We also use the English-French 20k sentence pairs from BTEC as a small scale experiment. As development and test sets, we the CSTAR03 and IWSLT04 held out sets, respectively.

Both languages were tokenized with Penn Tree Bank tokenizer\footnote{https://www.cis.upenn.edu/~treebank/tokenization.html}. As for casing, word-based models trained using the lowercased parallel data. At testing time, we uppercase using the model using the script at~\citep{moses}. This is a common practice for word-based models, as the sparcity induced by the same word in different casings (\examp{Play} vs \examp{play}) is problematic for word based models. For the character-based model, the model is trained on the true case on both source and target sides. That is, the model is responsible for interpreting and generating true cased sentences. Finally, evaluation is performed using BLEU~\citep{Papineni02bleu:a} and we always use a single sentence as reference. 

As for the hyper parameters of the model, all LSTM states and cells are set to 150 dimensions. Word projection dimensions for the source and target languages $d_{s,w}$ and $d_{t,w}$ were set to 50. Similarly, the character projection dimensions were also set to $d_{s,c}$ and $d_{t,c}$. For the alignment model, $d_{z}$ was set to 100. Finally, the beam search at the work level $k_w$ was set to 5, and the beam size for the character level search $k_c$ was set to 5. Training is performed until there is no BLEU~\citep{Papineni02bleu:a} improvement for 5 epochs on the development set. Systems are trained using mini-batch gradient descent with mini-batches of 40 sentence pairs. For word-based models on Europarl, rather than normalizing over the whole word inventory, we use noise contrastive estimation~\citep{GutmannHyvarinen:NoiseContrastiveEstimation}, which subsamples a set of 100 negative samples at each prediction. 

\subsection{Results}

For the BTEC dataset, the word-based neural model achieves a BLEU score of 15.38 in the French to English translation direction, while the character-based model achieves a score of 15.45. On the Europarl, the word-based model obtains a BLEU score of 19.29, while our proposed character-based model obtains a BLEU score of 19.49. While, differences are not significant, this is a strong results if we consider that previous work~\citep{Vilar:2007:WTL:1626355.1626360,neubig13mtj} revealed significantly lower translation quality using character-based approaches. 

\paragraph{Word Representation}

In order to analyse the word representation obtained after training the model, we generate the vector representation of the C2W models trained on the Europarl translation task with 600k sentence pairs for all words in the training set. Table\ref{top-5} provides examples for words in English and Portuguese, and words whose representations are closest to them, measure in terms of cosine similarity. We observe evidence that the fact that word representations are composed from characters as atomic units is not a limitation of their representative capabilities, as the model can learn that orthographically divergent words (\examp{answer} vs. \examp{response}) have similar meanings. Compared to the word-based model, we can observe that in many cases, the C2W model, prefers to gather words that are orthographically similar, such as different forms of \examp{answer} and \examp{responder}, which does not happen in word lookup tables. This is because, different conjugations in English are not always translated into the same word, and as lookup tables do not regard the orthographic similarity between words, source words that do not share translations are essentially distinct words and placed in different vector spaces. However, the sequential nature of the C2W model, makes it desirable to place these words in the same vector space as less memorization is required. Consequently, not only less parameters are needed to encode such word groups, but also, in the event a unknown verb conjugation is observed at test time, a reasonable vector can still be found for such a word. For instance, if we do not know the translation of \examp{answered}, we can still translate the word as \examp{answer} and obtain a reasonable translation of the sentence. 

Evidently, the assumption that similar words have similar meanings is not always true. For instance, the word \examp{well-founded} is not similar to \examp{much-needed}. In such cases, our model would behave equivalently to the word-based model.

\begin{table}
\begin{center}
\scalebox{0.80}{
\begin{tabular}{c|c|c|c|c|c}
\multicolumn{3}{c}{C2W Model} & \multicolumn{3}{|c}{Word Lookup Table}\\
\hline
\emph{answer}& \emph{well-founded} & \emph{responder} & \emph{answer} & \emph{well-founded} & \emph{responder}\\
\hline
response & well-balanced & respondeu & reply & described & reagir\\
answers & much-needed & responderam & response & impressed & aderir\\
reply & self-employed & responda & answering & bizarre & responda\\
answered & uncontrolled & responde & join & santer & aceder\\
replies & inherited & respondem & question & unclear & agradecer\\
\hline
\end{tabular}
}
\end{center}
\caption{\label{top-5} Most similar in-vocabulary words under the C2W model (left) and the word look up table (right); the two first query words are obtained from the source language projections (English) and the last word is obtained from target language projections (Portuguese).}
\end{table} 

\paragraph{Word Generation}
A strong aspect in the V2C model is that the model can generate unseen words. In Table~\ref{unk}, we provide three examples of unknown words that have been generated in Portuguese. The first is the translation of unknown word \examp{subsidisation} to \examp{subsidade}. Unfortunately, this is incorrect as \examp{subsidade} does not exist in the Portuguese vocabulary. However, it is encouraging to observe that the model is trying to translate the unknown word from its parts. Firstly, the English suffix \examp{-ation} and the Portuguese suffix \examp{-dade} are common endings for nouns. Furthermore, it can roughly copy the stem of the word \examp{subsi}. Unfortunately, the correct translation is \examp{subs\'{i}dio}. As the model has not seen any of these forms, it is unable to decide, which is the correct form. This hints that the V2C model could be pre-trained on large amounts of parallel data in order to recognize existing word forms, which will be left as future work. On the other hand, the word-based model simply translates this word to \examp{autores}, which is the translation for \examp{authors}. 

An example of an instance that a unseen word correctly is the plural of the noun \examp{reconstrução}. This is done by learning the general Portuguese rule for nouns ending in \examp{-ão}, which is converted into \examp{-ões}, whereas in general nouns are converted into plural by simply adding an \examp{-s}. The reason for generating the plural form, while the English word is in singular, is the fact that its preceding word \examp{as} is an determiner for plural words. This hints that cross word dependencies in the generation model are perserved.

\begin{table}
\begin{center}
\scalebox{0.80}{
\begin{tabular}{c|l}
\hline
original & $\ldots$ that does not mean that we want to bring an end to \textbf{subsidisation} .\\
character translation & $\ldots$ isso não significa a questão de que se trata de um fim à \textbf{subsidade} .\\
word translation & ... Isso não significa que isso , para conseguir reduzir os \textbf{autores} .\\  
%\hline
%original & ... between your \textbf{assertions} and reality . \\
%character translation & ... entre as suas \textbf{aceitações} e a realidade . \\
%word translation & ... entre as suas \textbf{opiniões} e a realidade .\\
\hline
original & the budget for the \textbf{reconstruction} of ...\\
character translation & o orçamento para as \textbf{reconstruções} ...\\
word translation & o orçamento inerente à \textbf{reconstrução} ...\\
\end{tabular}
}
\end{center}
\caption{\label{unk} Examples of unknown words that have been generated. The unknown word in the translation as well as their aligned words are marked in bold. The \examp{original}, \examp{character translation} and \examp{word translations} rows correspond to the original sentence, the translation by the character-based neural model and the translation obtained by the word-based model, respectively}
\end{table} 

\section{Related Work}

Our work is related to the recent advances in neural machine translation models, where a single neural network is trained to maximize the conditional probability of the target sentence $t$, given the source $s$. The different models proposed define different architectures to estimate this probability. These include the usage of convolutional archituctures~\citep{kalchbrenner13emnlp}, LSTM encoder-decoders~\citep{DBLP:journals/corr/SutskeverVL14} and attention-based models~\citep{DBLP:journals/corr/BahdanauCB14}. However, in all these approaches, the representation and generation of words are always performed at the word level, using a word lookup table and softmax. 

In our work, we focus on the definition of mechanisms to perform the word representation and generation on the character level. Thus, our methods are applicable to neural MT models proposed previously. On the representation level, it has been shown that words can be composed using character-based approaches~\citep{icml2014c2_santos14,DBLP:journals/corr/KimJSR15,ling:2015,miguel:2015}.

Generating output as a sequence of characters is receiving increasing attention in other domains. For example, speech recognition models~\citep{chan:2015,maas:2015} and language modeling~\citep{sutskever:2011,mikolov2012subword}. While this work is the first neural architecture we are aware of that uses hierarchical models, a variety of Bayesian language and translation models have been proposed that use subword  models to generate word types which are in turn used to generate text~\citep{chahuneau:2013,goldwater:2011}.

\section{Conclusion}

In this work, we presented an approach to perform automatic translation using characters as atomic units. Our approach uses compositional models to compose words from individual characters in order to learn orthographically sentient word representations. Then, we define a generation model to produce translated words at the character level. We show that our methods can improve over equivalent word-based neural translation models, as our models can learn to interpret and generate unseen words. 

As we present an end-to-end translation system that makes the open vocabulary assumption, we leave much room for future work, as our models make very simplistic assumptions about language. Much of the prior information regarding morphology~\citep{chahuneau13morphogen}, cognates~\citep{beinborn2013cognate} and rare word translation~\citep{DBLP:journals/corr/SennrichHB15} among others, should be incorporated for better translation.

\bibliography{iclr2016_conference}
\bibliographystyle{iclr2016_conference}

\end{document}